\documentclass{lincolncsuthesis}

\usepackage{hyperref}
\usepackage{svg}
\usepackage{amsmath}
\usepackage[frozencache=true,cachedir=minted-cache]{minted}

\title{\bfseries Generating Band-Limited Adversarial Surfaces Using Neural Networks}

\author{Roee Ben-Shlomo, Yevgeny Men}

\thesisDegree{BSc}

\thesisProgramme{Electrical Engineering}

\thesisSchool{EE department}

\thesisUniversity{Technion, Israel Institute of Technology}

\thesisStudentNumber{Supervised by Ido Imanuel}

\date{2021}

\thesisHeaderContents{CRML \& GiP Project Booklet}

\begin{document}

\maketitle



\begin{abstract}
    Generating adversarial examples is the art of creating a noise that is added to an input signal of a classifying neural network, and thus changing the network’s classification, while keeping the noise as tenuous as possible.
    
    While the subject is well-researched in the 2D regime, it is lagging behind in the 3D regime, i.e. attacking a classifying network that works on 3D point-clouds or meshes and, for example, classifies the pose of people’s 3D scans.
    
    As of now, the vast majority of papers that describe adversarial attacks in this regime work by methods of optimization. In this project we suggest a neural network that generates the attacks. This network utilizes PointNet's architecture with some alterations. 
    
    While the previous articles on which we based our work on have to optimize each shape separately, i.e. tailor an attack from scratch for each individual input without any learning, we attempt to create a unified model that can deduce the needed adversarial example with a single forward run.
    
    One could have a look at our \href{https://github.com/yevgm/DeepAdv3D}{\underline{GitHub repository}} which is publicly available.

\end{abstract}

\thesisTables
\thesisBodyStart


\chapter{Introduction}
In this section we will briefly introduce the various topics this project is about.

\section{Adversarial attacks}

Generating adversarial examples refers to the art of creating carefully perturbed input data that is meant to induce an alteration of the output predicted by the machine learning model, while keeping the noise as tenuous as possible, such that the data perturbation is undetectable to a human observer. 

\begin{figure}[htb]
    \centering
    \includegraphics[width=15cm]{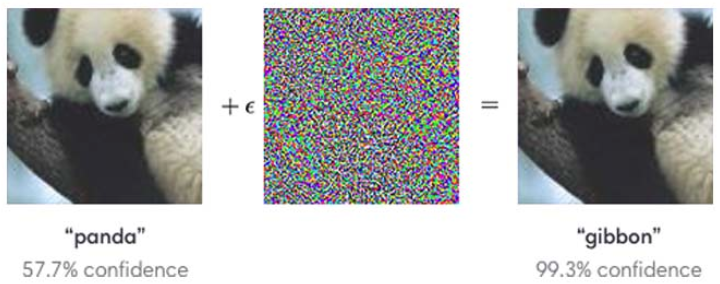}
    \caption{The 2D attack demonstrates transforming the image of a panda to a gibbon while making the images indistinguishable.}
    \label{fig:2d-adv-example}
\end{figure}

Creating an adversarial attack in the 2D domain refers to generating adversarial images as shown in Figure \ref{fig:2d-adv-example}.

\begin{figure}[htb]
    \centering
    \includegraphics[width=15cm]{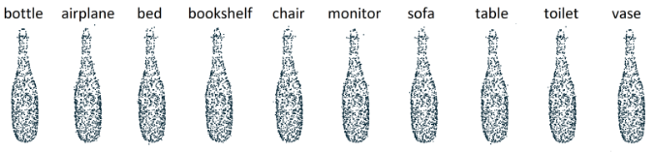}
    \caption{The 3D attack demonstrates transforming the bottle to whichever class in the data-set, while making the point clouds indistinguishable.}
    \label{fig:3d-adv-example}
\end{figure}

Adversarial attacks are also possible in the 3D domain, specifically targeting point clouds, as was shown by \cite{xiang2019generating} and demonstrated in Figure \ref{fig:3d-adv-example}. The exact data-set and classifier which were used in this project will be further elaborated in this booklet.

\section{FAUST dataset}
\begin{figure}[htb]
    \centering
    \includegraphics[width=15cm]{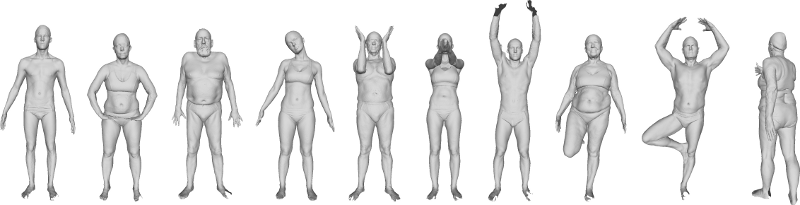}
    \caption{FAUST data-set.}
    \label{fig:faust}
\end{figure}

Created by \cite{Bogo:CVPR:2014}, FAUST is the primary\footnote{alongside SHREC-14, \cite{Pickup2014}.} data-set that was used in this project.
It includes 10 subjects, as shown in Figure \ref{fig:faust}, each performing 10 different poses. In this project we split the data as follows: the training set includes 70 of the meshes, the validation set 15 and the test set includes another 15.

\section{Background for the 3D domain}

In order to discuss the main article that we rely on, as well as our methodologies, we need to establish some background for the analysis of 3D shapes, as well as the basic definitions.

\subsection{3D shapes}
We model them as 2-Riemannian manifolds $\mathcal{X}$ embedded in $\mathbb{R}^3$, possibly with boundary $\partial \mathcal{X}$. We denote by $\mathcal{F}\left(\mathcal{X}\right)$ a Sobolev space of real-valued functions on $\mathcal{X}$, and use the inner product $\left\langle f,g\right\rangle =\int_{\mathcal{X}}f\left(x\right)g\left(x\right)\text{d\ensuremath{x}}$, where $\text{d}x$ is the standard volume form. To each shape $\mathcal{X}$ we attach the positive semi-definite Laplace-Beltrami operator $\triangle:\mathcal{F}\left(\mathcal{X}\right)\rightarrow\mathcal{F}\left(\mathcal{X}\right)$, which admits the spectral decomposition:

\begin{align}
  \triangle\phi_{i}\left(x\right)=\lambda_{i}\phi_{i}\left(x\right), && x\in\text{int}\left(\mathcal{X}\right) \\
  \left\langle \nabla\phi_{i}\left(x\right),\vec{n}\left(x\right)\right\rangle = 0, && x\in\partial\mathcal{X}
\end{align}

into eigenvalues $0 = \lambda_1 < \lambda_2 \leq \lambda_3 \leq \dots $, assumed to be ordered non-decreasingly, and the associated eigenfunctions $\phi_1, \phi_2, \phi_3, \dots$, which form an orthogonal basis for $\mathcal{F}(\mathcal{X})$. As shown above, we adopt homogeneous Neumann boundary conditions where $\vec{n}$ denotes the unit vector normal to the boundary.

The canonical ordering of the eigenvalues makes it so that truncating the Fourier-like series expansion of any scalar function $f \in \mathcal{F}(\mathcal{X})$ to the first $k$ terms: $$f\left(x\right)\approx\sum_{i=1}^{k}\left\langle \phi_{i},f\right\rangle \phi_{i}\left(x\right)$$ yields a \textit{band-limited} approximation of $f$ with bandwidth $k$. In fact, the orthogonal basis ${\phi_i}$ is optimal for approximating functions with bounded gradient magnitude in the $L_2$ sense, as described by \cite{aflalo2014optimality}.

\subsection{Discretization}

In the discrete setting, 3D shapes are sampled at $n$ points $x_1, \dots, x_n$
and approximated by a triangle mesh with vertex positions $\mathbf{X} \in \mathbb{R}^{n\times 3}$, and where each edge $e_{ij} \in E$ belongs to at most two triangle
faces $T_{ijk}$ and $T_{jih}$. Scalar functions $f$ are discretized as vectors
$\mathbf{f}\in\mathbb{R}^n$ with the values $f(x_i)$ for $i=1,\dots, n$, and linearly interpolated
within each triangle. Inner products $\left\langle f,g\right\rangle $ are discretized as $\mathbf{f}^\top \mathbf{Ag} $,
where $ \mathbf{A}$ is a $n\times n$ diagonal matrix of local area elements $a_{i}=\frac{1}{3}\sum_{jk:ijk\in T}A_{ijk}$ ($A_{ijk}$ is the area of triangle $T_{ijk}$). Vector fields $V: \mathcal{X} \rightarrow \mathbb{R}^3$ are discretized as matrices $\mathbf{V}\in\mathbb{R}^{n\times3}$, and their intergration $\left(\int_{\mathcal{X}} ||V(x)||^2_2 \text{d}x\right)^{0.5}$ is discretized as $ ||\mathbf{V}|| = \sqrt{\text{tr}(\mathbf{AVV^\top)}}$ .

Following linear FEM discretization, the Laplacian $\triangle$ is defined in terms of $\mathbf{A}$ and of a symmetric matrix $\mathbf{W}$ of edge weights:
$$
w_{ij}=\begin{cases}
-\frac{1}{2}\left(\text{cot}\alpha_{ij}+\cot\beta_{ij}\right) & ,e_{ij}\in E\\
\sum_{k\neq i}w_{ik} & ,i=j
\end{cases}
$$
Where $\alpha_{ij}, \beta_{ij}$ are the opposite angles to edge $e_{ij}$. A generalized eigenproblem $\mathbf{W\Phi} = \mathbf{A \Phi}\text{diag}(\lambda)$ is solved for computing the Laplacian eigenvalues (stored in vector $\mathbf{\lambda} \in \mathbb{R}^k$) and eigenvectors (stored column by column in the matrix $\mathbf{\Phi} \in \mathbb{R}^{n\times k}$). 

\chapter{Literature review}
Here we'll discuss the 3 main articles which the project relies on. The first one will lay the foundations and general algorithm for attacking in the 2D domain, while the other two will generalize and improve the method towards the 3D domain.

\section{Towards Evaluating the Robustness of Neural Networks, \cite{carlini2017evaluating}}

Carlini \& Wagner introduced in their paper 3 new 2D adversarial attack algorithms that are successful with 100\% probability. The initial formulation defines the problem of finding an adversarial instance for an image $x$ as follows:

$$
\begin{aligned}
& \text{minimize}
& & \mathcal{D}\left(x,x+\delta\right) \\
& \text{such that}
& & C\left(x+\delta\right) = t \\
&&& x+\delta\in \left[0,1\right]^n
\end{aligned}
$$

where $x$ is fixed, and the goal is to find $\delta$ that minimizes  $\mathcal{D}\left(x,x+ \delta \right)$ . That is, we want to find some small change $\delta$ that we can make to an image $x$ that will change its classification, but so that the result is still a valid image. Here  $\mathcal{D}$ is some distance metric; in the article $L_0$, $L_2$ \text{and} $L_\infty$ are used, but in the project only $L_2$ was used. When we discuss our project and the main article in the 3D domain we'll discuss why these metrics are only good for the 2D domain and fails in 3D.

The article doesn't use neural networks. in fact, there is no learning involved: an optimization method is used on each attack separately and nothing is being saved for later usage. 

The above formulation is difficult for existing algorithms to solve directly, and the constraint $C\left(x+\delta\right) = t$ is highly non-linear. Therefore, the article expresses it in a different form that is better suited for optimization. an objective function $f$ is defined such that $C\left(x+\delta\right) = t$ if and only if $f\left(x+\delta\right) \leq 0$. There are many possible choice of $f$, and the one that's important to us is:
$$f_{6}\left(x'\right)=\left(\max_{i\neq t}\left(Z\left(x'\right)_{i}\right)-Z\left(x'\right)_{t}\right)^{+}$$

Where $\left(e\right)^+$ is short-hand for $ \max{\left(e,0\right)}$ and $Z$ is the output of all layers of the neural net that defines our classifier, except for the softmax, so $Z\left(i\right)$ are the logits. 

Now, instead of formulating the problem as defined above, we use the alternative formulation:

$$
\begin{aligned}
& \text{minimize}
& & \mathcal{D}\left(x,x+\delta\right) + c \cdot f\left(x+\delta\right)\\
& \text{such that}
& & x + \delta \in \left[0,1\right]^n
\end{aligned}
$$

Where $c > 0$ is a suitably chosen constant. These two formulations are equivalent, in the sense that there exists $c > 0$ such that the optimal solution to the latter matches the optimal solution to the former. After instantiating metric $\mathcal{D}$ with an $l_p$ norm, the problem becomes: given $x$, find $\delta$ that solves:

$$
\begin{aligned}
& \text{minimize}
& & || \delta ||_p + c \cdot f\left(x+\delta\right)\\
& \text{such that}
& & x + \delta \in \left[0,1\right]^n
\end{aligned}
$$

We note that a constant with a similar purpose is later used in our project, referred to as \textit{reconstruction loss constant}.

\begin{figure}[htb]
    \centering
    \includegraphics[width=12cm]{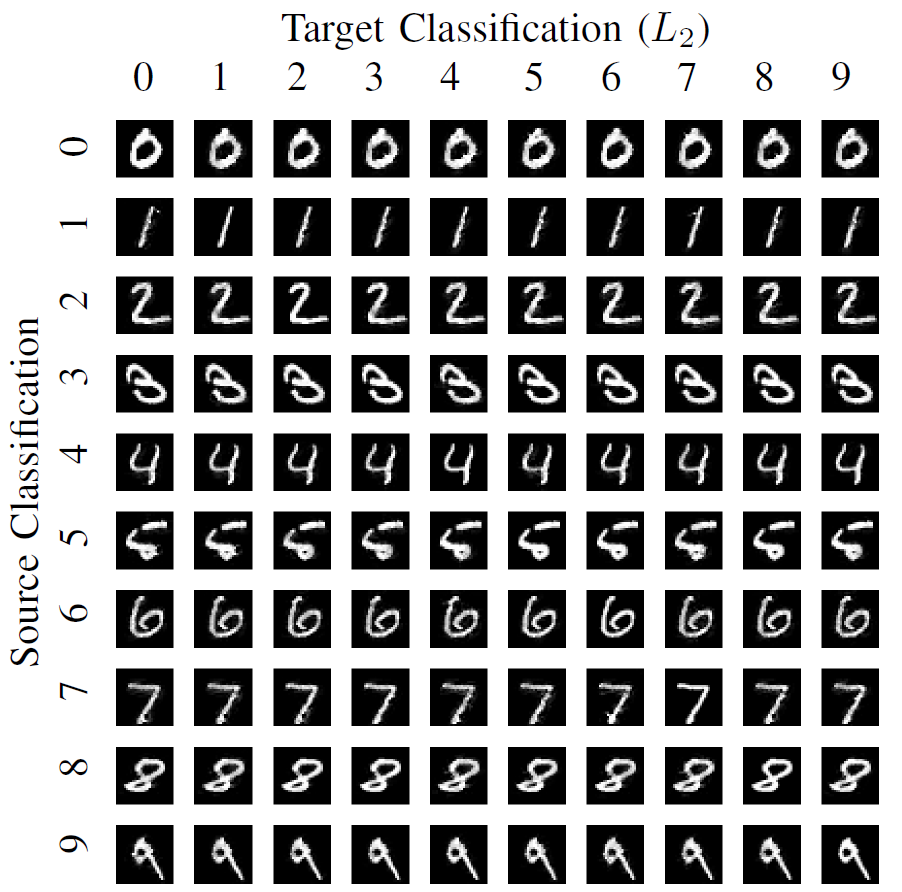}
    \caption{Carlini \& Wagner's attack performed on MNIST dataset.}
    \label{fig:carlini}
\end{figure}

The best way to choose $c$ is to use the smallest value of $c$ for which the resulting solution $x^*$ has $f\left(x^*\right) \leq 0$. This causes gradient descent to minimize both of the terms simultaneously instead of picking only one to optimize over first, and it's done by an exponential search.
As shown in Figure \ref{fig:carlini}, when the method is applied to the MNIST dataset, Carlini \& Wagner managed to create an attack for every source/target pair.

\section{PointNet, \cite{qi2017pointnet}}

\begin{figure}[htb]
    \centering
    \includegraphics[width=12cm]{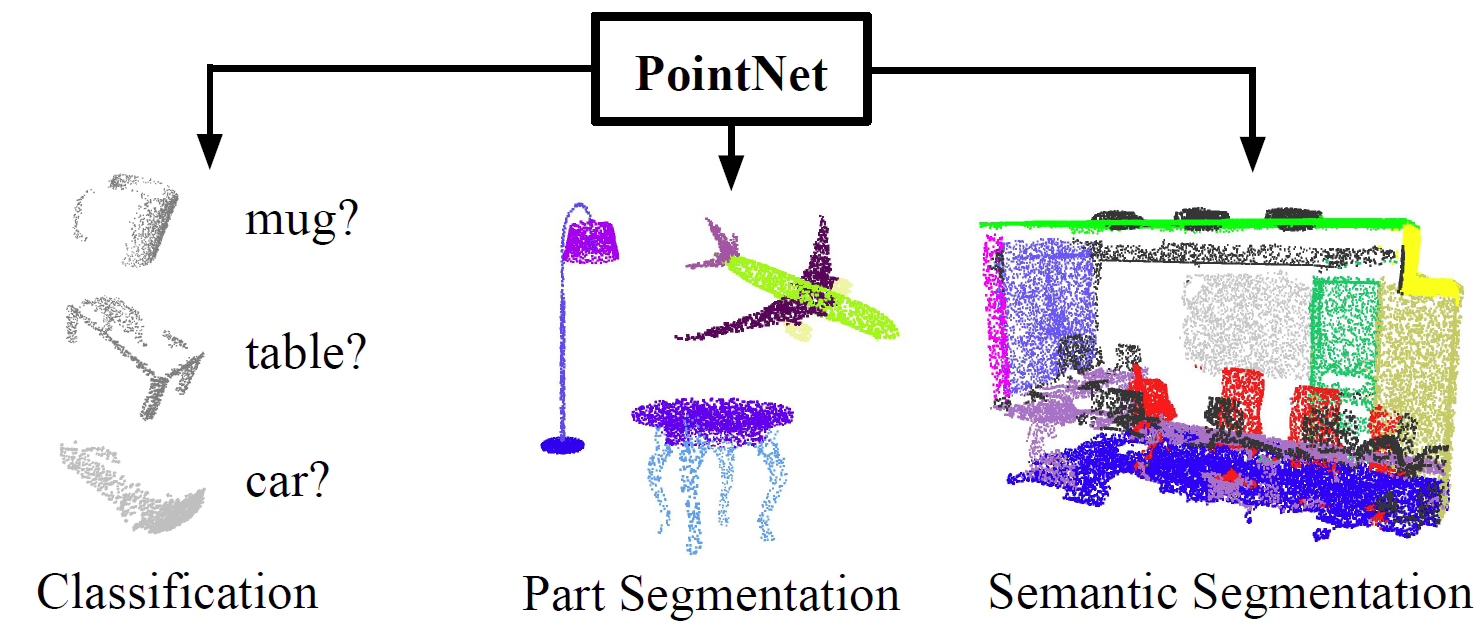}
    \caption{Applications of PointNet. We'll be using the left one.}
    \label{fig:pointnet-applications}
\end{figure}

This paper makes the transition from the 2D domain to the 3D domain such as a point cloud, which is an important type of geometric data structure. Qi et al. designed a novel type of neural network that directly consume point clouds, which well respect the permutation invariance of points in the input. As shown in Figure \ref{fig:pointnet-applications}, it provides a unified architecture for applications ranging from object classification, part segmentation, to scene semantic parsing.

\begin{figure}[htb]
    \centering
    \includegraphics[width=17cm]{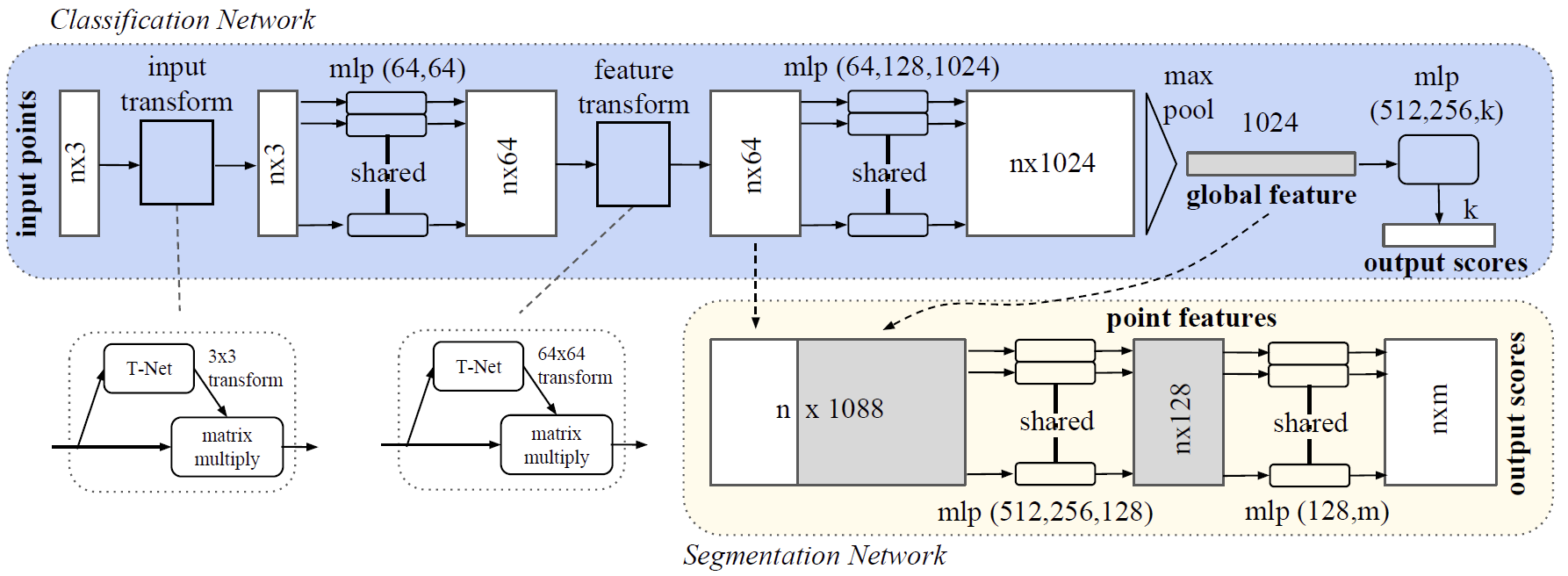}
    \caption{PointNet Architecture}
    \label{fig:pointnet-architecture}
\end{figure}

A point cloud is represented as a set of 3D points $\left\{  P_i | i=1,...,n\right\}$, where each point $P_i$ is a vector of its $\left(x,y,z\right)$. The network, with an architecture that's shown in Figure \ref{fig:pointnet-architecture}, outputs $k$ scores for all the $k$ candidate classes, while keeping in mind 3 main properties that the point cloud has to maintain:
\begin{enumerate}
    \item Unordered. Unlike pixel arrays in images or voxel arrays in volumetric grids, point cloud is a set of points without a specific order. In order words, a network consumes $N$ 3D point sets needs to be invariant to $N!$ permutations of the points provided, where $N$ is the number of points.
    \item Interaction among points. The points are from a metric space. It means that points are not
    isolated, and neighboring points form a meaningful
    subset. Therefore, the model needs to be able to
    capture local structures from nearby points, and the
    combinatorial interactions among local structures. It's worth noting that while it's indeed a property that PointNet tries to work by, only the later work PointNet++ by \cite{qi2017pointnetpp} improves on this matter.
    \item Invariance under transformations. As a geometric
    object, the learned representation of the point set
    should be invariant to certain transformations. For
    example, rotating and translating points all together
    should not modify the global point cloud category nor
    the segmentation of the points.
\end{enumerate}

As shown in Figure \ref{fig:pointnet-architecture}, the classification network takes $n$ points as input, applies input and feature transformations, and then
aggregates point features by max pooling. The output is classification scores for k classes.

\section{3D Adversarial Attacks, \cite{10.1111:cgf.14083}}

This is the main paper on which we relied. It was published in 2020 and claims that while increasing attention has been placed on the image domain, the study of adversarial perturbations for geometric data has been notably lagging behind. Mariani et al. shows that effective adversarial attacks can be concocted for surfaces embedded in 3D, under weak smoothness assumptions on the perceptibility of the attack. They address the case of deformable 3D shapes in particular, and introduces a general model that we'll shortly discuss. 

Just as the previous article, the objective here is that given an input $\mathbf{X}$, generate a new adversarial shape $\mathbf{X'}$ such that:
$$ C(\mathbf{X'}) \neq C^* (\mathbf{X}) \quad \text{and} \quad \mathbf{X'} \sim \mathbf{X}$$
Where $C^*(\mathbf{X'})$ denotes the ground-truth label of  $\mathbf{X}$ and $\sim$ signifies that $ \mathbf{X'}$ is imperceptibly close to $ \mathbf{X}$ according to some metric. In Carlini's article we discuss the $L_2$ metric, but we'll shortly understand why we need a different metric here, as well as in our project.

\begin{figure}[htb]
    \centering
    \includegraphics[width=15cm]{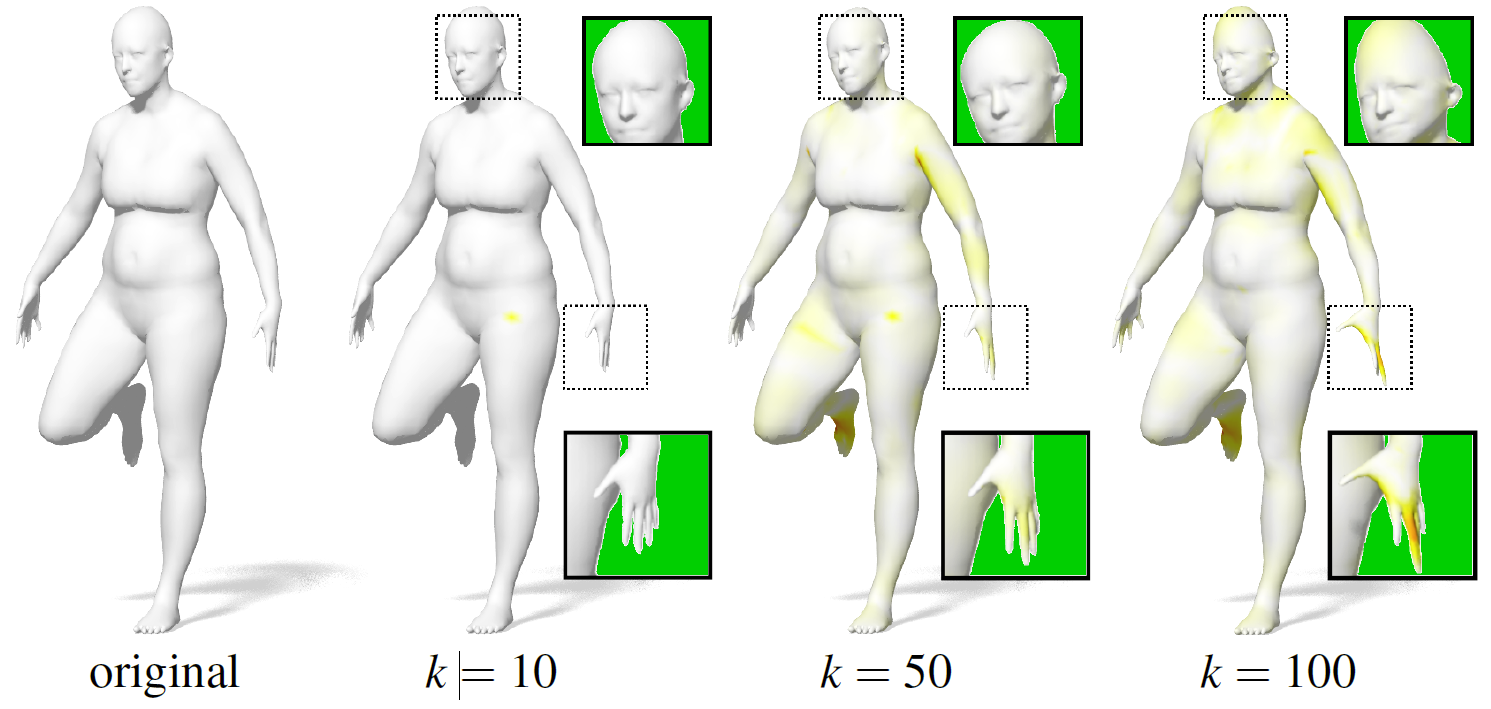}
    \caption{Increasing the spectral bandwidth $k$ of the perturbation. For lower $k$ values the shape isn't misclassified.} 
    \label{fig:3d-bandwidth}
\end{figure}

The paper models the adversarial shape $\mathbf{X'}$ as a perturbation of $\mathbf{X}$ along a deformation field $\mathbf{V} \in \mathbb{R}^{n\times 3}$:
$$ \mathbf{X'} = \mathbf{X} + \mathbf{V} $$
Having a small norm for the vector field $\mathbf{V}$ is not enough for the attack that we're looking for: it also has to be \textit{smooth} . Smooth deformations preserve local neighborhoods, and prevent the formation of adversarial jittering that is observed with point cloud attacks.
This desired smoothness is enforced by passing to a subspace parameterization:
$$ \mathbf{V} = \mathbf{\Phi v} $$
Where $\mathbf{\Phi}$ contains the first $k$ Laplacian eigenvectors of $\mathbf{X}$, and $v \in \mathbb{R}^{k\times 3}$ is a set of expansion coefficients representing $\mathbf{V}$ in the reduced basis. With this parameterization, smoothness is easily controlled by varying spectral bandwidth $k$, as illustrated by Figure \ref{fig:3d-bandwidth}. For large $k$, one admits high-frequency oscillations in the deformation field, while for small $k$ we only retain the smoother, low-frequency behavior. 
They emphasize that we require smoothness for the deformation field $\mathbf{V}$ only, and not for the entire embedding $\mathbf{X'}$ , which would instead lead to an undesirable loss of geometric detail on the surface. 
In the chapter that describes our project we shall see all of these "mistakes" take place, as we've had lots of trials and different methods. 

The model that creates the attack is based on Carlini's article: we require $\mathbf{X'}$ to minimize the penatly function:
$$
h_t(\mathbf{X'}) = \max{\{Z_i': i \neq t\}} - Z_t'
$$
In particular, $h_t(\mathbf{X'})<0$ if and only if the constraint that $C(\mathbf{X'}) =t$ holds exactly.

Minimizing this equation alone would lead to $\mathbf{X'}$  deforming arbitrarily. Therefore, we pass to the unconstrained optimization problem:
$$
\min_{v\in \mathbb{R}^k} ||\mathbf{\Phi v}|| + c h_t(\mathbf{X+\Phi v})^+
$$
Where $a^+ = \max{\{0,a\}}$, and the constant $c$ is similar to the constant from Carlini's article. Just like Carlini's paper, an exponential search for the smallest value of $c$ is proposed.

In practice, we replace the point-wise measure $ || \mathbf{\Phi v}||$ with the pair-wise distortion:

$$
\sum_{i=1}^{n}\sum_{j\in\text{NN}\left(i\right)}\left(||\mathbf{X}_{i:}-\mathbf{X}_{j:}||-||\mathbf{X}_{i:}'-\mathbf{X}_{j:}'||\right)^{2}
$$

Where $\mathbf{X' = X+ \Phi v}$ and $\mathbf{X}_{i:}$ denotes the $i$-th row of matrix $\mathbf{X}$. This promotes local euclidean distances to be preserved in an as-rigid-as-possible fashion. In the next chapters it's shown that this metric plays a significant role in our project.

\chapter{Methodology}
In this chapter we'll present the novelty of our project, as well as our methodologies, different models, loss functions and more.

While the 2 papers that we introduced in the literature review chapter achieved nice looking attacks with high success rates, they used solely optimization methods. This is where our project gets in, as we suggest a setting that utilizes neural networks to create the attacks.

\section{Formal Problem Statement}
For a given 3D dataset $X$ we would like to create, with a neural network, a new dataset $X'$ such that:

\begin{enumerate}
    \item $X'$ is visually similar to $X$.
    \item While $X$ is classified correctly by some classifier, $X'$ is misclassified. 
\end{enumerate}

In our case, the data-set is FAUST and the classifier is a PointNet classifier; both were discussed in the introduction.
The classifier was trained by us and got classifications percentage of 90\%, 87\% and 87\% on the train, validation and test sets accordingly. During the classifier training we fed it with augmented shapes, i.e we translated and rotated the shapes, otherwise during the adversarial training the attacking network would exploit this weakness and create "attacks" that aren't meaningful.

\section{Models Architecture}
We altered PointNet's architecture to create an auto-encoder as shown in the following two subsections.
This auto-encoder was altered in order to create 2 different models:

\subsection{Model 1}
\begin{figure}[htb]
    \centering
    \includegraphics[width=17cm]{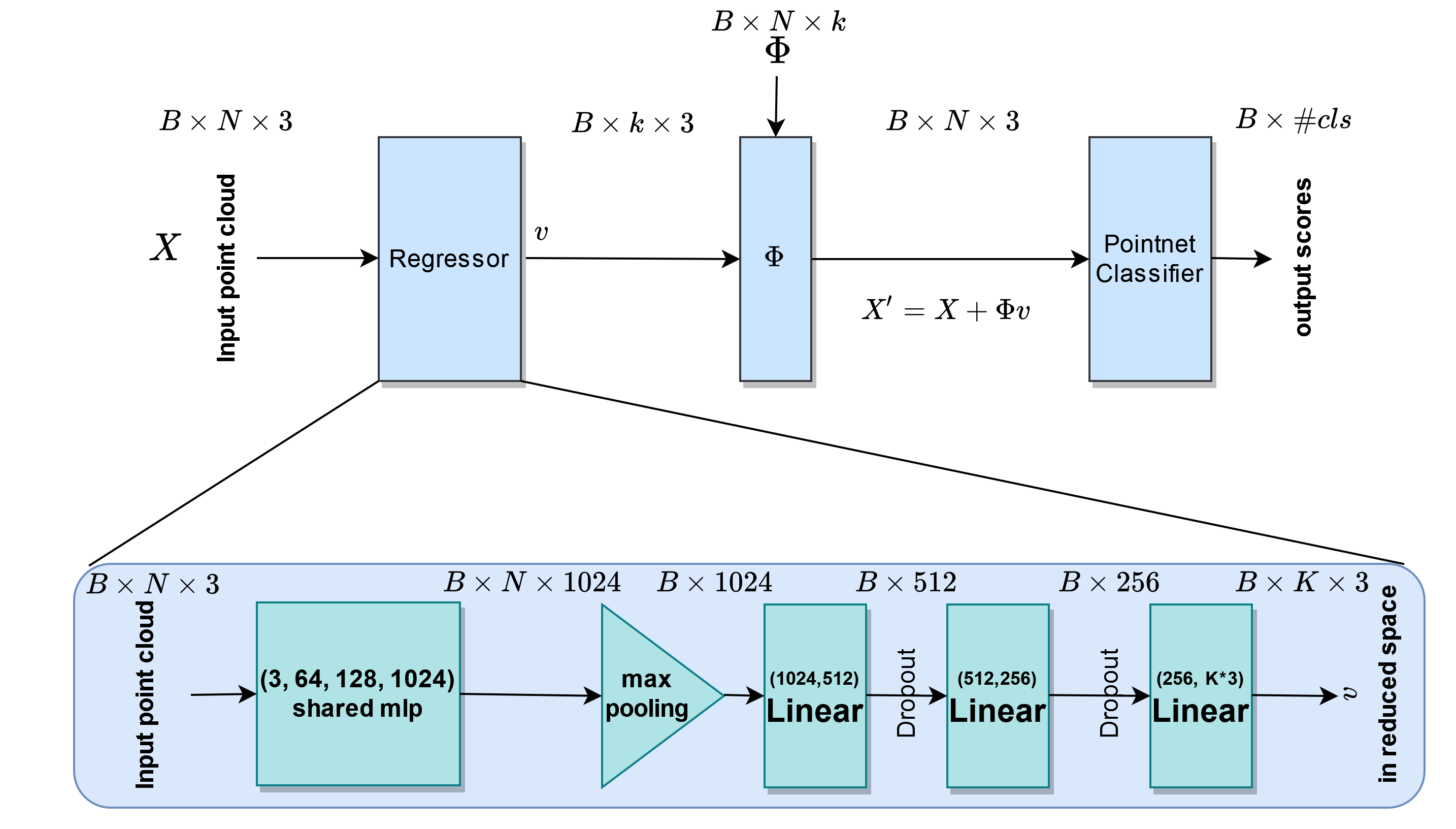}
    \caption{Model 1 architecture. It's based on PointNet's architecture in a form of a regressor and outputs a perturbation $v$ in the reduced space. The perturbation is then multiplied by the eigenvalues matrix $/Phi$ and added to the original shape to create the attack, which is passed to the trained PointNet classifier.}
    \label{fig:model1}
\end{figure}

As shown in Figure \ref{fig:model1}, this model uses the central article's method in the neural setting: regressing for the optimal smooth deformation field while making sure to attain a smooth, natural-looking example.

\subsection{Model 2}

\begin{figure}[htb]
    \centering
    \includegraphics[width=17cm]{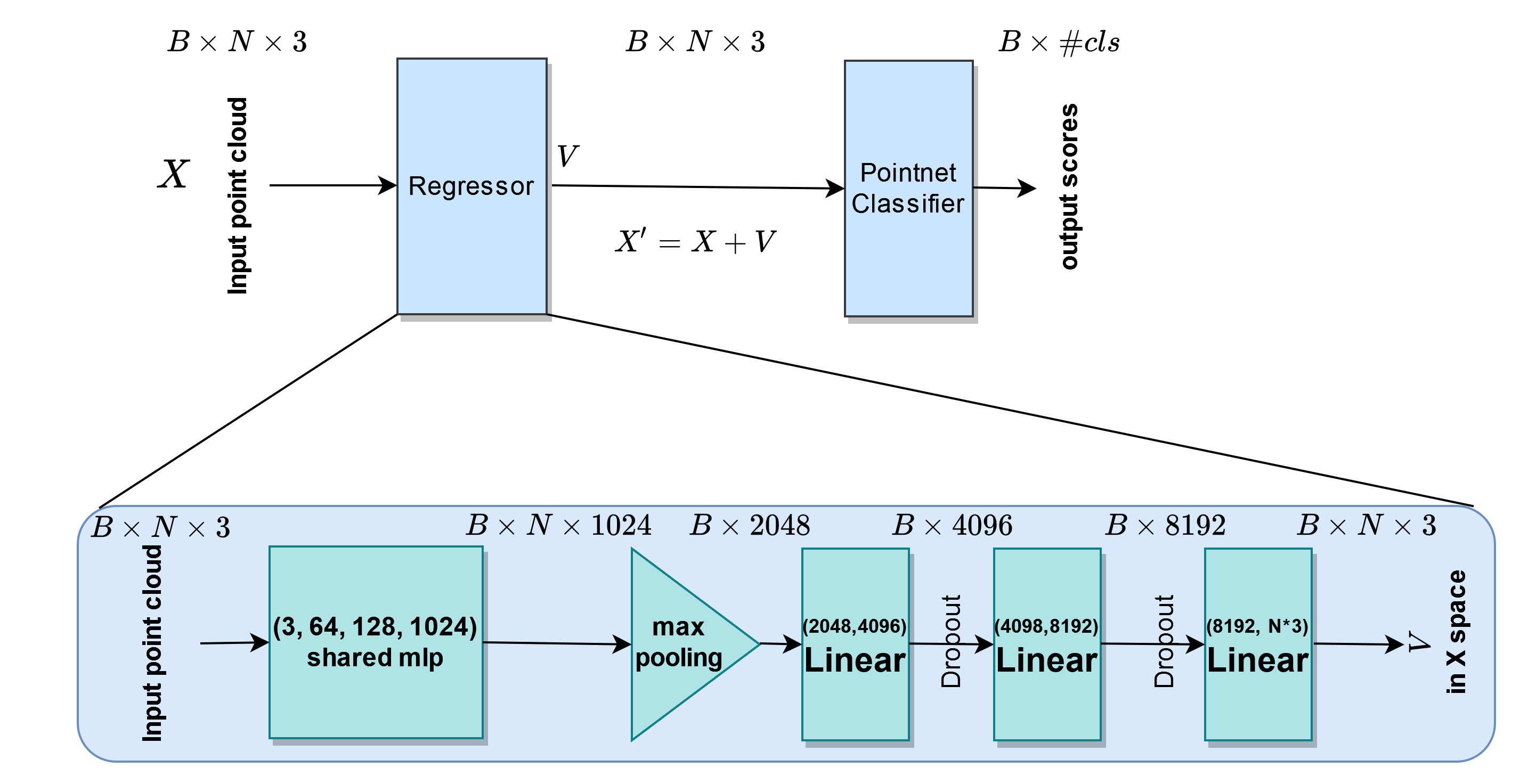}
    \caption{Model 2 architecture. This model tries to have a direct inference of the target shape. It directly extracts the perturbation vector field $\mathbf{V}$ without using the eigenvectors' decomposition: $X' = X + V$.}
    \label{fig:model2}
\end{figure}

We'll see in the next section that this model turned out to produce unnatural looking attacks, as the smoothness of the perturbation is not regularized.

\section{Metrics and Loss Functions}

Both of the models in the previous subsections outputs a matrix of shape $B \times \text{\#classes}$: for each input we fed the PointNet classifier with our adversarial attack that was created by the regressor, and thus we have the probability that each of the attacks belongs to each of the classes (10 in the case of FAUST). 

After receiving this matrix, we had numerous tries of loss functions that were used, all of which shared the same basic principle that relies on the discussed articles:
$$
L = L_\text{Misclass} + c\cdot L_\text{Recon}
$$
The misclassification loss remained the one that was proposed by the main article:
$$
(\max{\{Z_i': i \neq t\}} - Z_t')^+
$$
Where, as mentioned in the literature review, $Z_i$ are the logits of the PointNet classifier, $a^+ = \max{\{a,0\}}$ and $h_t(\mathbf{X'}) = \max{\{Z_i': i \neq t\}} - Z_t$ is only negative when a misclassification occurs. 

As for the reconstruction loss, we suggest several similarity metrics. The results section exhibits results of different mixtures of them:

\begin{enumerate}
    \item $L_2: ||\mathbf{X'}-\mathbf{X}||_2$
    \item Edge Loss: $\frac{1}{\#E}\cdot\sum_{\left(i,j\right)\in E}\left|\frac{||V_{i}'-V_{j}'||}{||V_{i}-V_{j}||}-1\right|$
    \item Local Euclidean: $\sum_{i=1}^{n}\sum_{j\in\text{NN}\left(i\right)}\left(||\mathbf{X}_{i:}-\mathbf{X}_{j:}||-||\mathbf{X}_{i:}'-\mathbf{X}_{j:}'||\right)^{2}$
    \item Chamfer Distance: $\sum_{i\in\mathbf{X'}} d_X(i)$
\end{enumerate}

Where $d_X(i)$ is the minimal distance between vertex $i$ in $\mathbf{X'}$ and some vertex in $\mathbf{X}$.
The edge loss penalizes edges of the mesh that changes their length from the original input to the adversarial attack, or in other words, this loss preserves edge lengths.
The local euclidean, taken from the main article, captures local neighborhoods of vertices and preserves as-rigid-as-possible similarities.

In addition to the mentioned reconstruction losses, we applied in some of the runs a Laplacian Smoothing: $\frac{1}{N}\sum_{i=1}^{B}\#\mathbf{V}_{i}||\mathbf{LV}_{i}||_{2}^{2}$ . The expression $L
\mathbf{V}$ is an approximation of the mean curvature of the input shape, which is exactly what we were trying to minimize since some of the attacks were having displacements of single vertices. This phenomenon resembles spikes, and throughout this booklet shapes with these spikes will be referred to as spiky.

\chapter{Results}
\begin{figure}[h]
    \centering
       \subfloat[Misclassfication Percentage]{%
          \includegraphics[width=9cm, height=6cm]{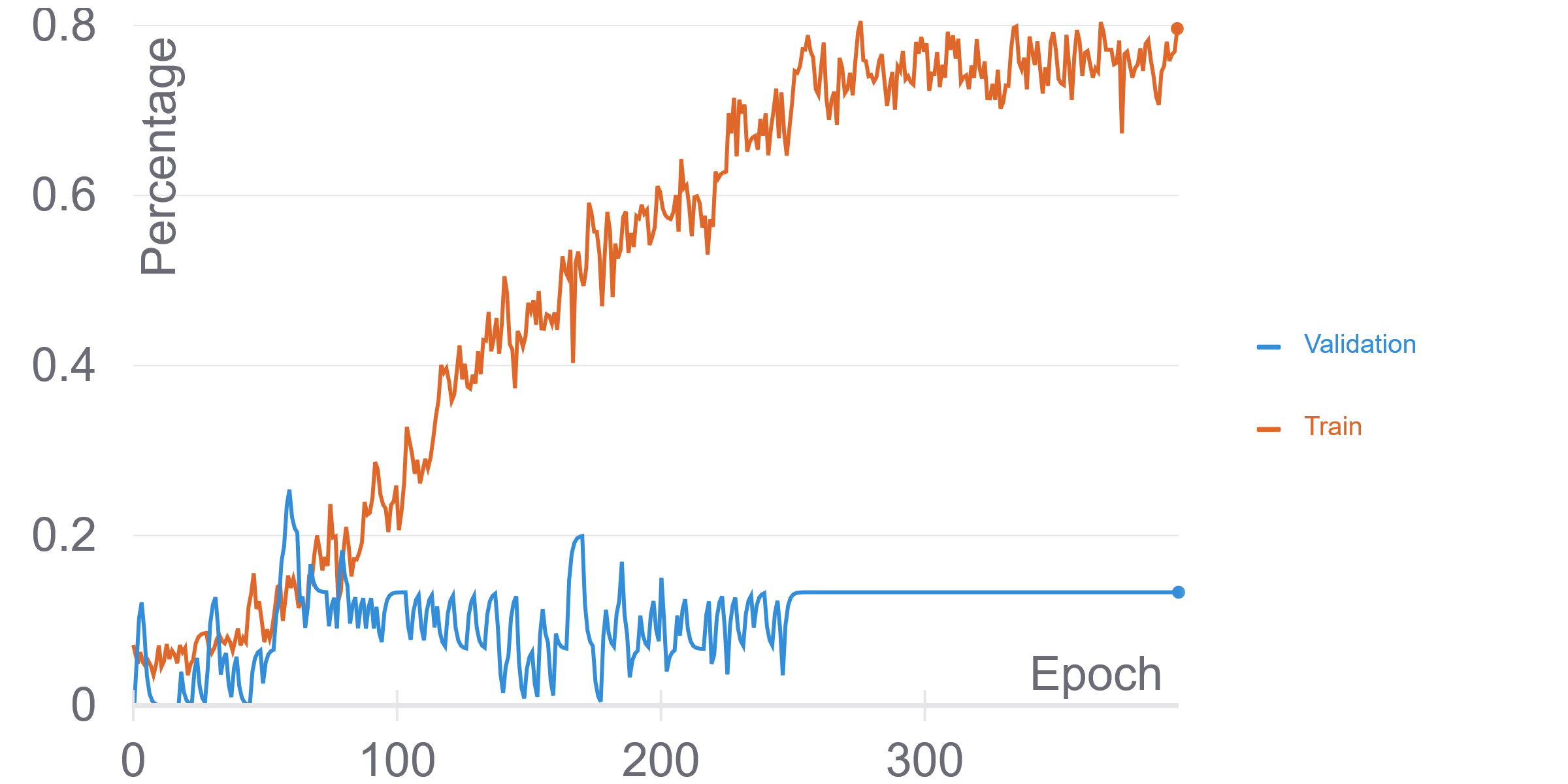}%
          \label{fig:left1}%
       } 
       \subfloat[Misclassification Loss]{%
          \includegraphics[width=9cm, height=6cm]{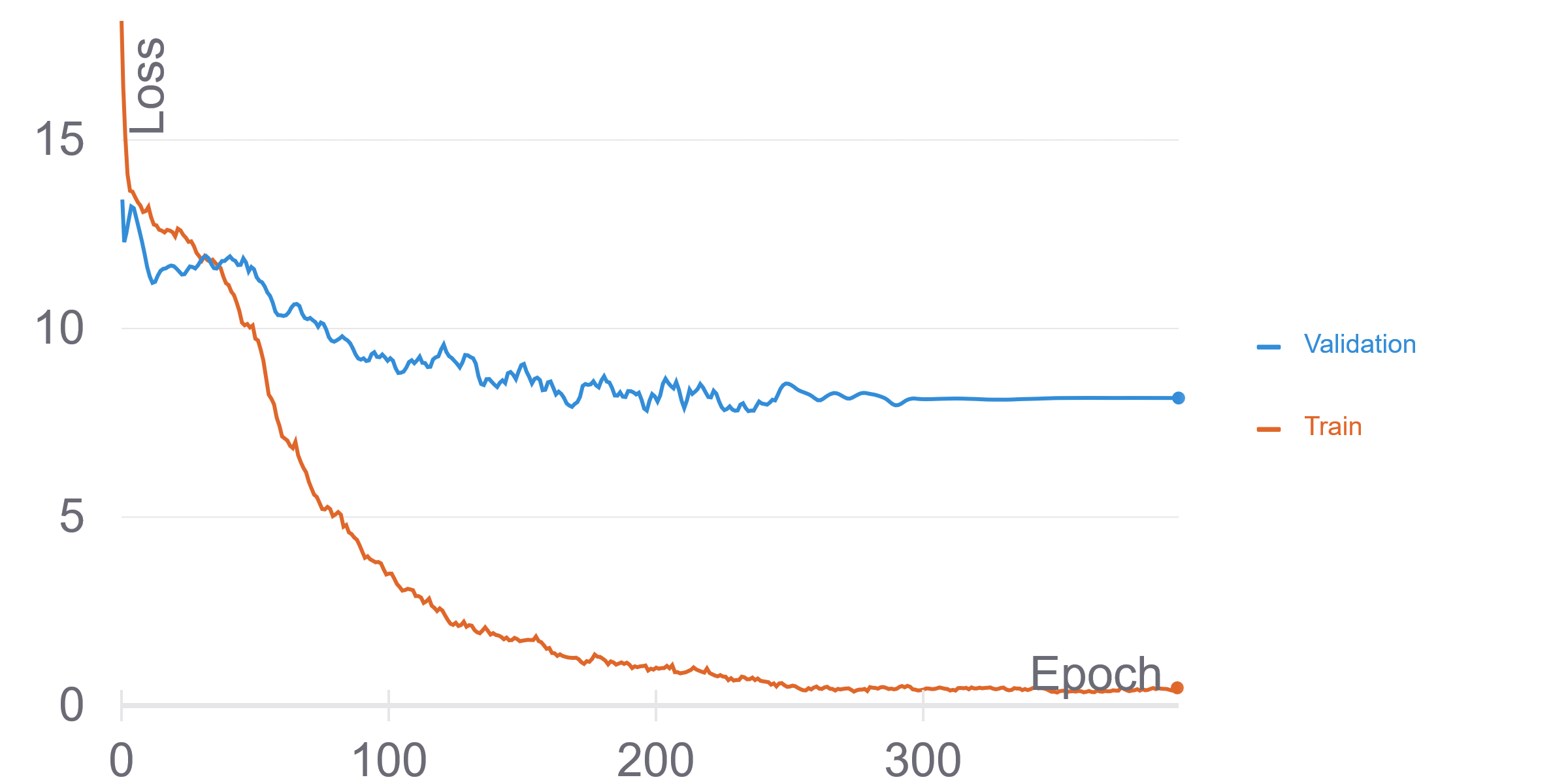}%
          \label{fig:right1}%
       }
       
       \caption{Misclassification metrics of one of our best runs, picked from WandB.}
       \label{fig:misclass-graphs}
\end{figure}

\begin{figure}[H]
    
    \centering
       \subfloat[Reconstruction Loss]{%
          \includegraphics[width=9cm, height=6cm]{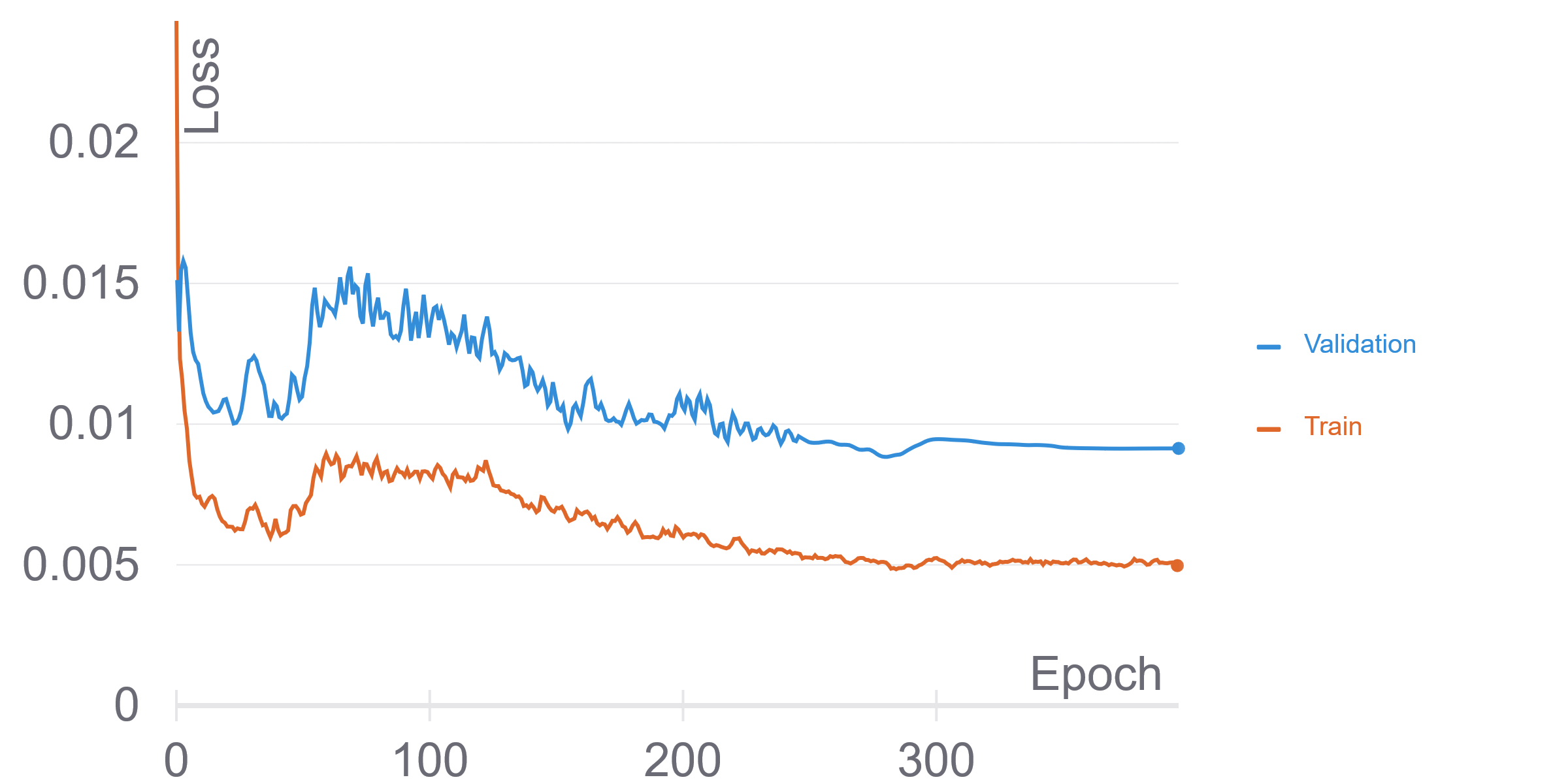}%
          \label{fig:left2}%
       } 
       \subfloat[Total Loss]{%
          \includegraphics[width=9cm, height=6cm]{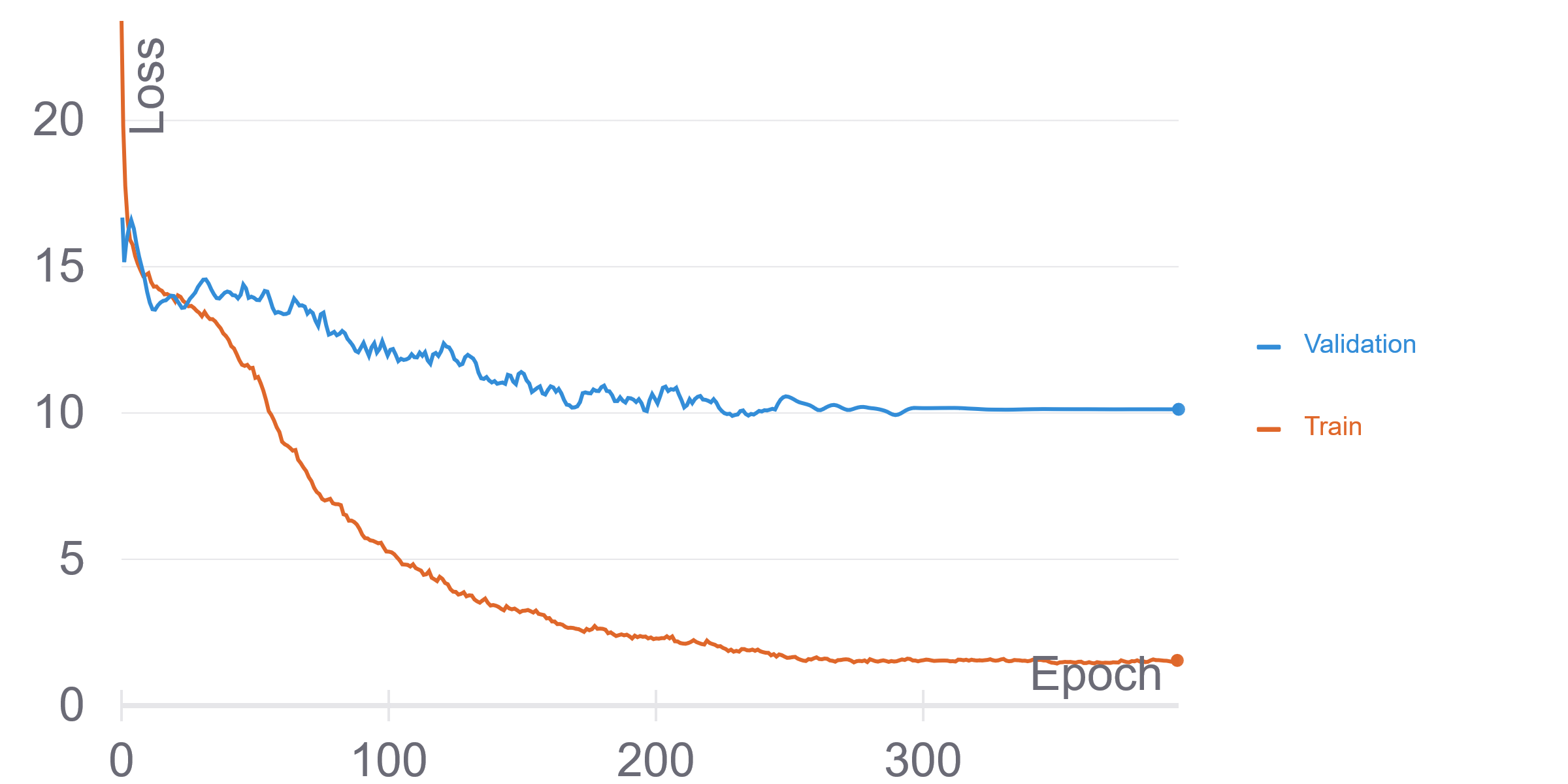}%
          \label{fig:right2}%
       }
       
       \caption{Reconstruction \& total losses of one of our best runs, picked from WandB.
       While the total loss of both training and validation set decreases steadily, only the training set exhibits a significant amount of reduction. In fact, Figure \ref{fig:misclass-graphs} shows that while the validation misclassification value gets lower than the initial value in Figure \ref{fig:right1}, Figure \ref{fig:left1} shows that we manage to misclassify mainly the training set, and we does not in fact manage to generalize to the validation set. This result will be discussed in the next chapter.}
       \label{fig:recon-total-graphs}
\end{figure}

In order to test the different methodologies we used Weights \& Biases' experiment tracking tool as well as their hyperparameter sweeps. In total over than 7,000 runs were made in order to get the best out of the models. Figures \ref{fig:misclass-graphs} and \ref{fig:recon-total-graphs} exhibits one of the runs that brought the best results in terms of misclassifications amount and most natural looking examples. 

\captionsetup*[subfigure]{position=bottom}

\captionsetup*[subfigure]{position=bottom}

\begin{figure}[htb]
    \centering
    \includegraphics[width=18cm]{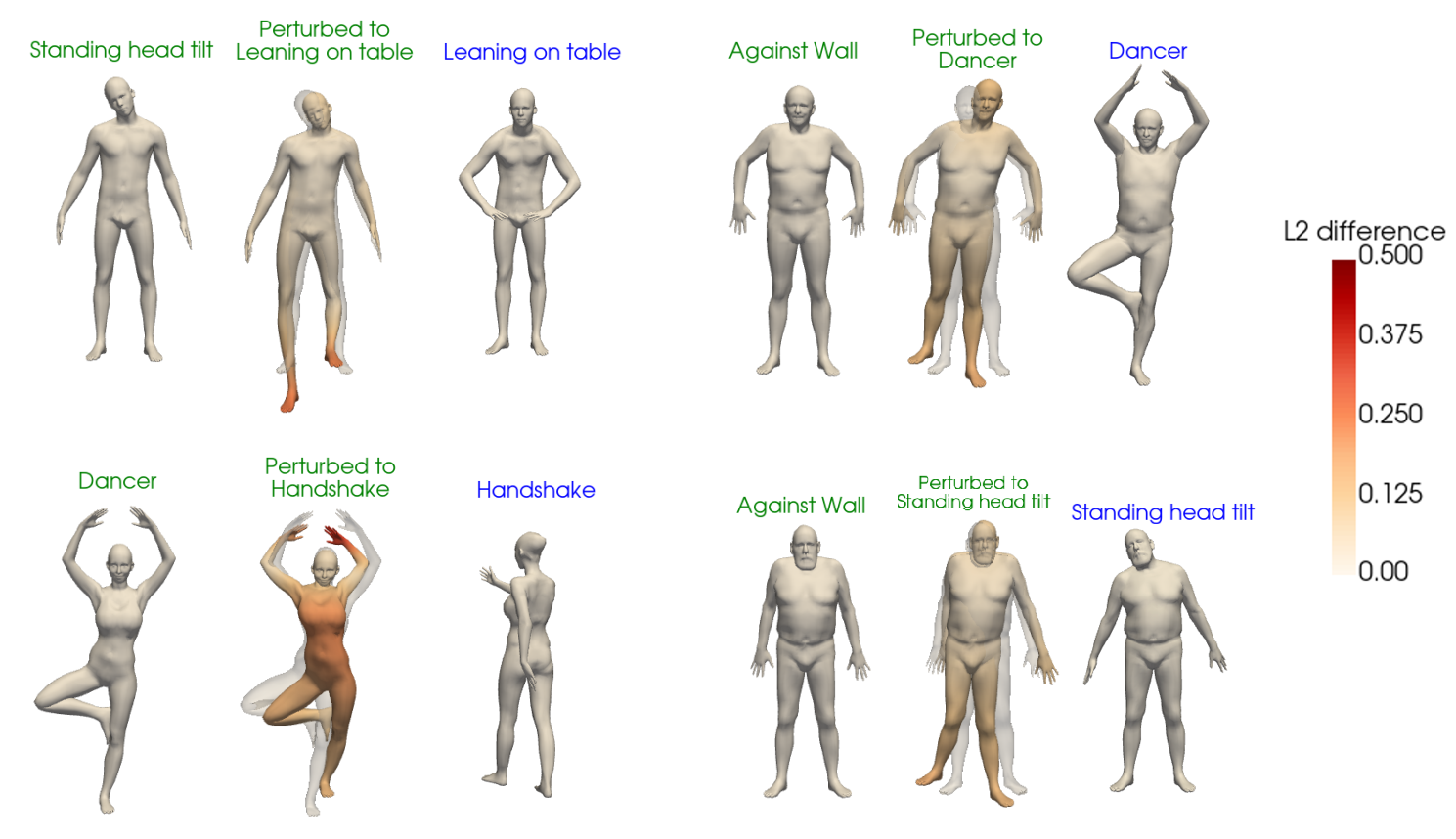}
    \caption{Some of the successful attacks that were produced with model \#1. It's reconstruction loss is local euclidean alone. From left to right in each attack: original input shape, adversarial attack on top of original input shape, target shape.}
    \label{fig:successful-attacks}
\end{figure}

\begin{figure}[htb]
    \centering
    \includegraphics[width=15cm]{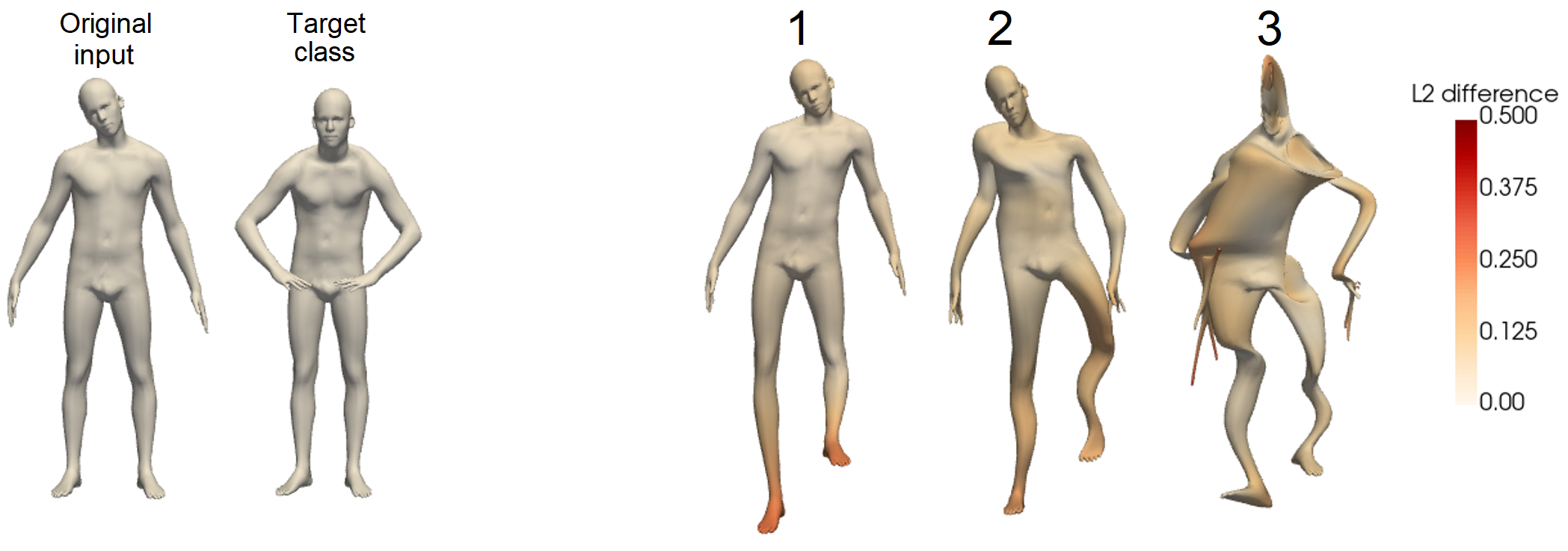}
    \caption{Comparing different reconstruction losses with Model \#1.}
    \label{fig:row1}
\end{figure}

\begin{figure}[htb]
    \centering
    \includegraphics[width=15cm]{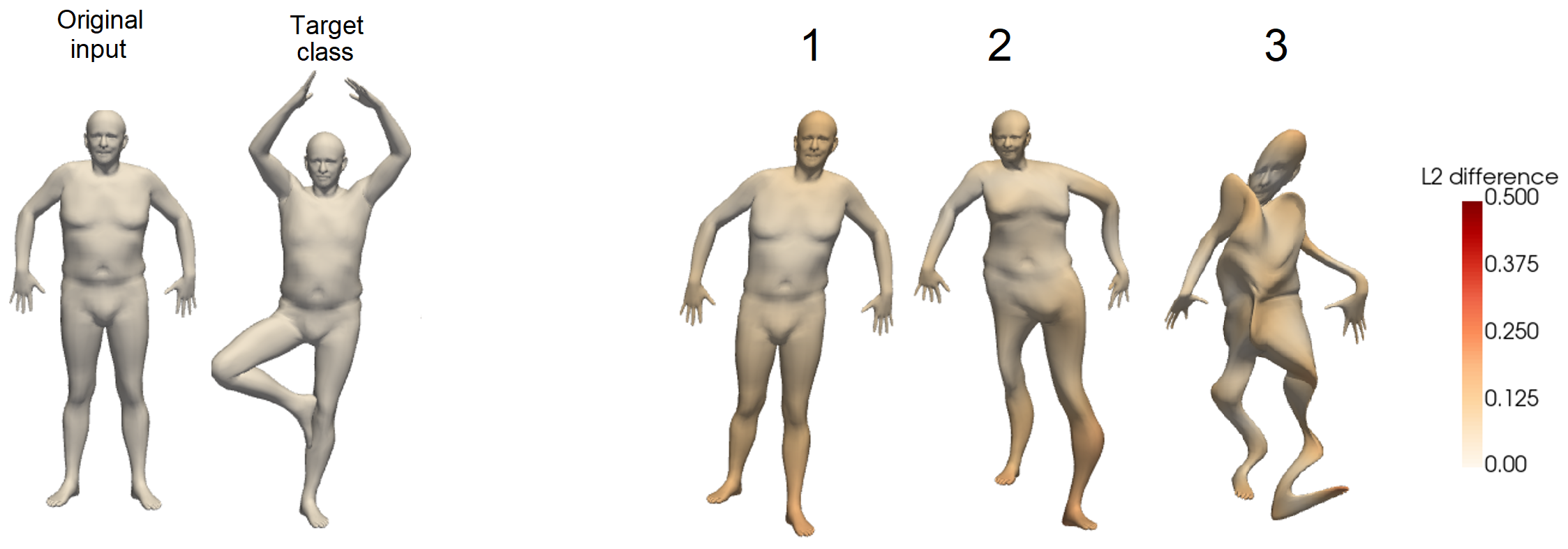}
    \caption{Some more model \#1 attacks using different reconstruction losses}
    \label{fig:row2}
\end{figure}

Figures \ref{fig:row1} and \ref{fig:row2} demonstrates the results of Model \#1 across multiple trials when it comes to the reconstruction loss. From left to right:
\begin{enumerate}
    \item The run that features the graphs in Figures \ref{fig:misclass-graphs}, \ref{fig:recon-total-graphs} and \ref{fig:successful-attacks} with local euclidean alone
    \item Local euclidean + centering the classifier's inputs
    \item Local euclidean + $L_2$ 
\end{enumerate}

\begin{figure}[htb]
    \centering
    \includegraphics[width=15cm]{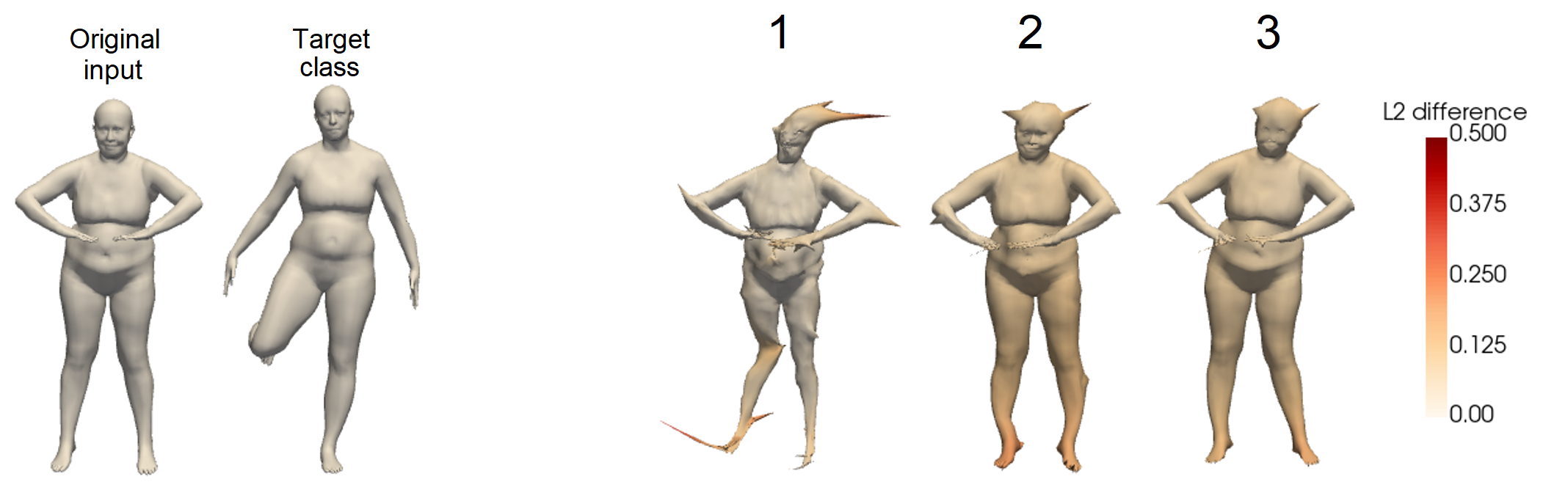}
    \caption{Some model \#2 attacks using different reconstruction losses}
    \label{fig:model2row1}
\end{figure}

Following model \#1, we show in Figure \ref{fig:model2row1} some attacks of model \#2. Once again, the misclassification loss remains constant among the trials, and the difference remains in the pick of reconstruction losses. From left to right:

\begin{enumerate}
    \item Local euclidean + edge loss
    \item Local euclidean alone
    \item Local euclidean + Laplacian smoothing
\end{enumerate}

\begin{table}[htb]
\centering
\begin{tabular}{l|l|l|l}
\textbf{Run type} & \textbf{Mean Curvature distortion} & \textbf{Edge loss}         & \textbf{$L2$} \\
\hline\hline
Run \#1         & $4.02$,    $3.51$,  $3.96$       & 4e-4, 4.5e-4, 4.5e-4   & $0.156$, $0.146$, $0.171$ \\
Run \#2         & $5.60$,   $5.07$,   $5.14$       & 4.2e-4, 4.7e-4, 4.6e-4 & $0.177$, $0.171$, $0.193$ \\
Run \#3         & $26.52$,  $22.87$,   $24.85$    & 7e-4, 7e-4, 7.2e-4    & $0.113$, $0.108$, $0.115$ \\
\hline
\cite{10.1111:cgf.14083} & $3.05$ & - & 0.062 \\
\end{tabular}
\caption{A comparison between the different types of runs of model 1. \textbf{From left to right: train, validation, test results}.
Each metric represents a measure of noticeability of the attack. Mean curvature distortion is the $L2$ difference between the original shape's and the adversarial example's mean curvature.}
\label{tbl:model1_table}
\end{table}

\begin{table}[htp]
\centering
\begin{tabular}{l|l|l|l}
\textbf{Run type} & \textbf{Mean Curvature distortion} & \textbf{Edge loss}         & \textbf{$L2$} \\
\hline\hline
Run \#1         & $31.23$,    $27.16$,  $30.21$       & 3.6e-4, 4e-4, 3.9e-4   & $0.051$, $0.053$, $0.053$ \\
Run \#2         & $10.73$,   $9.67$,   $10.11$       & 4e-4, 4.4e-4, 4.3e-4 & $0.061$, $0.061$, $0.068$ \\
Run \#3         & $11.45$,  $10.47$,   $11.17$    & 3.6e-4, 4e-4, 3.9e-4    & $0.052$, $0.051$, $0.062$ \\
\hline
\cite{10.1111:cgf.14083} & $3.05$ & - & 0.062 \\
\end{tabular}
\caption{A comparison between the different types of runs of model 2. Same as \ref{tbl:model1_table} but for model 2.}
\label{tbl:model2_table}
\end{table}

\clearpage
Table \ref{tbl:model1_table} compares the averaged results of model 1 between the different runs that are shown in Figures \ref{fig:successful-attacks}, \ref{fig:row1} and \ref{fig:row2}. While those figures present a clear visualization that shows that run 1 is better than run 2, which is better than run 3, this table allows to significantly compare different runs. 

Indeed, one can see that the table supports the visualizations. Taken from \cite{10.1111:cgf.14083}, we calculated the absolute curvature distortion metric, which is the absolute difference between the mean curvatures of our attack and the original shape, and averaged it over the train, validation and test sets. Our best run got a curvature distortion of 4.02, 3.51 and 3.96 on the train, validation and test sets accordingly, comparing to Mariani's 3.05.

A similar table, Table \ref{tbl:model1_table}, was made, and one can easily see that it also supports the visual results: the metrics aren't close to Mariani's metrics.

In addition to the mean curvature distortion, edge loss and $L_2$ were compared between the runs in the sake of completeness, as some of the runs were optimized by them. However, one can notice that they don't provide as good of a comparison as the mean curvature distortion. For example, run 3 of model 1 shows results that are nowhere close to natural, but only the mean curvature distortion shows a clear surge in comparison to run 1, while the edge loss doesn't show as much of it, and the $L_2$ metric even shows a better value for run 3.

\chapter{Discussion and Conclusions}
\begin{figure}[htb]
    \centering
    \includegraphics[width=18cm]{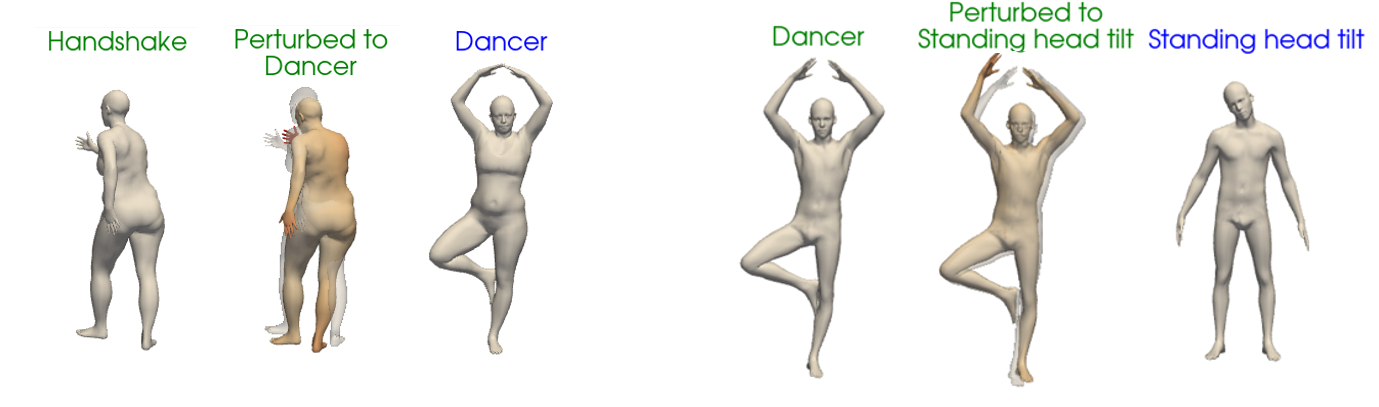}
    \caption{Some of the attacks that we generated after implementing Mariani's paper.}
    \label{fig:mariani-attacks}
\end{figure}

We started this project by implementing the main article by \cite{10.1111:cgf.14083}. 2 of the attacks that we managed to produce are presented in Figure \ref{fig:mariani-attacks}. Generally speaking, the attacks were looking natural and close to the inputs. In addition, using their methodologies we were able to generate attacks that were almost always successful (but not always natural looking), just as the article boasts.

We then proceeded to create model 1 as described in the previous chapters. The results appeared extremely unnatural and resembled at best the attack from run 3 in Figure \ref{fig:row1}. We adjusted the network's width and length, as well as the dropout layers, but couldn't manage to fix the problem.

At this point the simpler model, model 2, came in our mind, as we figured that something that's involved with the Laplacian had some sort of bug. After implementing model 2, the attacks were still very much distorted, this time spiky as well since there's no regularization on the smoothness of the perturbation, as shown in Figure \ref{fig:model2row1}. These "spikes" brought us to browse different papers across the topic of 3D meshes in order to find a metric that suits out problem and smooths out the attacks. This problem arises from the fact that our data is 3-dimensional, as well as having faces, in contrast to the simple $L_2$ metric that is sufficient in the 2D domain.

Looking for the bug, we tried overfitting our model to the training data, since generally every network should be able to achieve it. However, no matter how many different hyperparameter configurations we tried, we couldn't manage to get an overfit. After countless different checks, reading of our code and even observing the network's gradients during training using Weights and Biases, we solved the bug by removing the vast majority of our code and model and starting over, and managed to get an overfit. At this point we also had all of the different similarity measurements that we found.

From here after the bug was solved we continued with model 2. As shown in Figure \ref{fig:model2row1} we tried 3 different reconstruction losses: local euclidean + edge loss, local euclidean alone and local euclidean + Laplacian smoothing. All of the runs contained the local euclidean loss as any run without it turned out to be extremely spiky. However, this metric itself didn't suffice, as attack \#2 shows in Figure \ref{fig:model2row1}. We then tried adding edge loss, which tries to preserve the lengths of the edges in comparison to the original input (attack \#1). After seeing the results we moved on to add Laplacian smoothing in addition to the local euclidean. This mixture of losses indeed smoothed out the attacks, but we couldn't satisfy an equilibrium between misclassifying and producing natural-looking attacks, as only one of the two was obtainable.

Despite having better results than our runs that occurred before the fix of the bug, all of the mentioned runs with model 2 were producing attacks that looked unnatural. Hence, we moved on to add the laplacian, i.e. we created model 1. Right from the start we noticed that the strength of this model comes from the fact that no matter what hyperparameters or reconstruction metrics we used, the attacks turned out smooth. This comes from the fact that the perturbation is now projected on the Laplacian eigenspace. From the previous runs we had all of the similarity measurements ready, so all we had to do is run a couple of hyperparameter sweeps, one for each loss configurations.

These sweeps generated about 6,000 runs, and for each sweep, i.e. for different reconstruction loss configuration, we chose the best run to display. This was done by looking at the couple of runs that achieved the highest training misclassification number, as well as low reconstruction loss, and then picking the run whose attacks looked most natural and close to the input shapes. As shown in the previous chapter in the discussion about Figure \ref{fig:misclass-graphs}, those runs that managed to generate adversarial attacks that are visually similar to the original shapes\footnote{Some runs did in fact manage to generalize and produce about 50\% misclassifications on the validation set, but the attacks were too distorted to count as meaningful.} couldn't manage to generalize and misclassify the validation and test sets. In other words, over-fitting is evident and we weren't able to overcome this issue that was previously desired, even with different regularizations.

Having that said, at this stage we managed to achieve smooth and natural-looking attacks, as model \#1 does indeed exhibits the ability to successfully produce adversarial attacks on 80\% of the set of examples that its given. In this project what we were hoping for is the ability to create a generalization with a unified model that can deduce the needed adversarial example with a single forward run. That is a much stronger demand than both the articles we relied on, which base their attack methodology on direct iterative optimization. 

Moreover, as discussed in the literature review section, both of the base articles use an exponential search to find the optimal reconstruction constant $c$. In other words, for each different input they're able to meticulously fit its own constant, and not only that, but the constant that produces the best looking attack. In contrast, we used a single constant for the entire training set, and thus created a much more general methodology.



\printReferences


\end{document}